\documentclass{article}
\usepackage{spconf,amsmath,graphicx}
\usepackage{booktabs}
\usepackage{algorithm}
\usepackage{algpseudocode}
\usepackage[pagebackref,breaklinks,colorlinks]{hyperref}
\usepackage{subcaption} 
\usepackage{cite}
\usepackage{multirow}
\usepackage{booktabs}
\usepackage{tabularx}
\usepackage{graphicx}
\usepackage[table]{xcolor}
\usepackage[most]{tcolorbox}
\usepackage{pifont}
\usepackage{graphicx}
\usepackage{wrapfig}
\usepackage{colortbl}

\definecolor{Sijia_color}{rgb}{0.858, 0.188, 0.478}

\title{The Power of Few: Accelerating and Enhancing \\ Data Reweighting with Coreset Selection}
\name{Mohammad Jafari $^{1,}$\sthanks{The work was carried out during his (remote) summer internship at the OPTML Lab of Michigan State University.}  ~~~ Yimeng Zhang $^2$ ~~~ Yihua Zhang $^2$ ~~ Sijia Liu $^2$}
\address{$^1$ Sharif University of Technology, Tehran, Iran ~~~ $^2$ Michigan State University, MI, USA}
\begin{document}
%
\maketitle
\begin{abstract}
As machine learning tasks continue to evolve, the trend has been to gather larger datasets and train increasingly larger models. While this has led to advancements in accuracy, it has also escalated computational costs to unsustainable levels. Addressing this, our work aims to strike a delicate balance between computational efficiency and model accuracy, a persisting challenge in the field. We introduce a novel method that employs core subset selection for reweighting, effectively optimizing both computational time and model performance. By focusing on a strategically selected coreset, our approach offers a robust representation, as it efficiently minimizes the influence of outliers. The re-calibrated weights are then mapped back to and propagated across the entire dataset. Our experimental results substantiate the effectiveness of this approach, underscoring its potential as a scalable and precise solution for model training.
\end{abstract}
\section{Introduction}
\label{sec:intro}
In the evolving field of machine learning, numerous strategies have been developed to tackle the complexities of model training \cite{NIPS2012_c399862d,chen2023deepzero,zhang2024revisiting,zhang2022robustify,zhang2023text,zhang2023data,han2020tensor,zhang2020video,jia2022robustness,zhang2022advancing,zhang2022distributed,zhang2023missing,zhang2024introduction,zhang2022fairness,hou2022textgrad,zhuang2023pilot,khanduri2023linearly,zhang2023generate,zhang2024unlearncanvas,zhang2023robust,zhang2022revisiting,chen2022quarantine,chen2023understanding,fan2023salun,zhang2023visual}. Two particularly noteworthy techniques are coreset selection\cite{chen2012supersamples, gal2017deep, shen2018deep, shen2018deep, toneva2019empirical, borsos2020coresets, coleman2020selection, killamsetty2021glister, killamsetty2021retrieve, paul2023deep, xia2023moderate, jia2023robustness,zhang2024selectivity}, aimed at improving computational efficiency by selecting most representative training data or removing harmful training data, and data reweighting 
\cite{fan2018learning, petrović2020fair, wu2018learning, jiang2019identifying, zhao2019metricoptimized, 
lahoti2020fairness, ren2019learning, 
saxena2019data,
vyas2020learning, lin2018focal, 
dong2017class,
shu2019meta}, 
designed for generalization or faster convergence. While each has contributed to advancements in machine learning, they come with their own sets of challenges. Coreset selection, for instance, can sometimes overlook the nuanced diversity inherent in larger datasets. On the other hand, data reweighting incurs a significant computational burden, particularly when applied to expansive datasets.

\begin{figure}[t]
  \begin{center}
    \includegraphics[width=\linewidth]{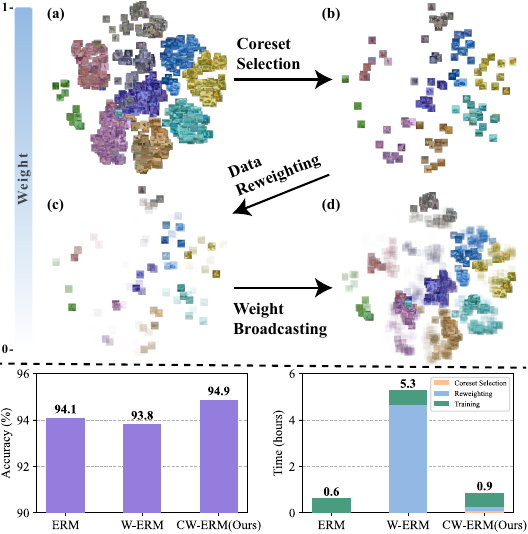}
    \vspace*{-6.5mm}
   \caption{ \footnotesize{Analysis of model performance and computational efficiency. The first part illustrates the three-stage process of our method: coreset selection, data reweighting, and weight broadcasting, visualized through four 2D t-SNE embeddings indicated by a, b, c, d. The second part compares the accuracy and time consumption of ERM, W-ERM, and our method (CW-ERM). Our approach not only yields the highest accuracy but also maintains a balance between computational efficiency and performance.}
    }
    \label{fig:Motivation}
  \end{center}
   \vspace*{-3em}
\end{figure}

Against this backdrop, we introduce a novel methodology that capitalizes on the advantages of both coreset selection and data reweighting for model training. As shown in \textbf{Fig. \ref{fig:Motivation}}, our proposed method not only addresses the limitations of each individual method but also synergistically enhances their strengths. By focusing on reweighting a strategically chosen coreset, we achieve computational efficiency without compromising on trained model performance.
The distinction of our proposed method lies in its dual commitment to both computational speed and robust performance, which is achieved by marrying the efficiency gains from coreset selection with the precision enhancements afforded by data reweighting. 
As a result, our proposed method not only quickens the pace of machine learning tasks but also elevates their accuracy, making it a compelling alternative to existing techniques.
We summarize the main contributions of this paper as follows:

\noindent \ding{172} (Methodology-wise) We propose a new framework that 
integrates coreset selection and data reweighting, striking a balance between training efficiency and model efficacy.

\noindent \ding{173} (Efficiency-wise) We unveil that less than $1\%$ of the dataset is sufficient for effective reweighting, dramatically increasing the efficiency of the process. 

\noindent \ding{174} (Performance-wise) We empirically validate our approach through comprehensive experiments on the CIFAR-10 and CIFAR-100 datasets, showing that our method maintains competitive levels of accuracy.

\section{Related Work}
\label{sec:background}
\textbf{Coreset selection.}
Coreset selection revolves around the idea of distilling a large dataset into a smaller, yet highly informative subset for efficient model training. Various strategies have been employed to achieve this, ranging from geometry-based methods like Herding \cite{chen2012supersamples} and K-Center \cite{sener2018active}, to uncertainty-based \cite{shen2018deep,gal2017deep,coleman2020selection} and error-based approaches \cite{paul2023deep, toneva2019empirical}. In geometry-based methods, samples are chosen based on their geometric attributes in the feature space \cite{chen2012supersamples, sener2018active}. For instance, the K-Center method minimizes the maximum distance from any data point to its nearest center in the selected subset. On the other hand, moderate coreset targets data points that are close to the median distance in the feature space \cite{xia2023moderate}.
Uncertainty-based methods aim to identify challenging samples that lie near the decision boundary or that the model is least confident about \cite{coleman2020selection, shen2018deep, gal2017deep}. Error-based methods such as GraNd and EL2N focus on samples that contribute most to the model's loss\cite{paul2023deep}. Additionally, some works have explored the use of bilevel optimization \cite{borsos2020coresets, killamsetty2021glister, killamsetty2021retrieve}, though these face scalability issues with large datasets. While some existing coreset selection methods may be computationally efficient, they often cannot match the performance achieved by training on the entire dataset.
\\ \\
\noindent\textbf{Data reweighting.}
Data Reweighting techniques often determine weights iteratively, either based on the training loss \cite{fan2018learning, petrović2020fair, wu2018learning} or by focusing on fairness discrepancies within the training data \cite{jiang2019identifying}. These strategies have been further developed by employing functions, typically parameterized, that map inputs directly to weights \cite{zhao2019metricoptimized, lahoti2020fairness}. Some methods even treat these weights as directly learnable variables \cite{ren2019learning}. However, these techniques generally do not produce a global set of weights, limiting their utility for tasks like post-training data compression. Previous approaches to data reweighting have targeted various objectives such as enhancing generalization, boosting resistance to noisy labels \cite{ren2019learning, saxena2019data, shu2019meta, vyas2020learning}, and addressing class imbalance issues \cite{lin2018focal, dong2017class}. Another set of works aims at reducing training time and achieving faster convergence by employing a learning curriculum tailored to individual instances\cite{saxena2019data}. Current data reweighting techniques are often not scalable and add computational overhead, making them less practical for large datasets.

\section{Problem Statement}

\textbf{Setup.}
This paper elucidates the terminological framework and overarching methodology underpinning the research. The study is situated in the domain of supervised learning, specifically targeting classification tasks. The primary dataset, denoted by \( \mathcal{S} \), comprises \( n \) samples, each characterized as \( (\mathbf{x}_i, y_i) \), where \( \mathbf{x}_i \) represents the image and \( y_i \) the corresponding label of the \(i\)-th sample. Central to the study is the concept of a \textit{coreset}, defined as a compact yet representative subset \( \mathcal{C} \subseteq \mathcal{S} \) of the original dataset. The procedure of \textit{Data Reweighting} entails the recalibration of sample weights \( w^c \) within this coreset. The research builds upon the foundational framework of Empirical Risk Minimization (ERM), designed to minimize the mean loss over the training dataset. Formally, ERM is articulated as:
\begin{equation}
\min_{\theta} \frac{1}{n} \sum_{i=1}^{n} \mathcal{L}(f(\mathbf{x}_i; \theta), y_i),
\end{equation}
\label{eq: loss}
where \( \mathcal{L} \) denotes the loss function and \( \theta \) the model parameters. An extension of ERM termed Weighted-ERM (W-ERM), incorporates sample weights \( \boldsymbol w \) into the loss function:
\begin{equation}
\min_{\theta} \frac{1}{n} \sum_{i=1}^{n} \boldsymbol w_i \mathcal{L}(f(\mathbf{x}_i; \theta), y_i),
\end{equation}
\label{eq: weighted_loss}

\noindent\textbf{Challenges.}
Contemporary data reweighting methodologies are confronted with significant computational challenges, especially when applied to the vast and intricate datasets prevalent in modern settings. Such inefficiencies critically impede their suitability for practical machine learning assignments which necessitate a harmony of speed and precision. Present techniques are further hampered by the absence of a cohesive framework that adeptly amalgamates the collective benefits of coreset selection and data reweighting. This gap results in a piecemeal landscape, punctuated with potential unexploited avenues for optimization. Consequently, two paramount \underline{challenges} emerge:
\ding{172} the distillation of datasets into concise, yet informative coresets for efficient data reweighting;
\ding{173} the broadcasting of coreset data weights to the whole dataset without undermining the efficacy of the model.
By addressing these challenges, our work aims to bridge the gap between effectiveness and efficiency - enabling scalable, precise, and fast machine learning through the synthesis of coreset selection and data reweighting. We propose a unified framework that harnesses the strengths of both techniques to advance the state-of-the-art.

\section{Methods}
\label{sec:method}

In this section, we elucidate our research methodology, which seamlessly integrates three pivotal components: coreset selection, coreset reweighting, and the broadcasting of recalibrated weights back to the full dataset. Our proposed method, Coreset Reweighting for ERM (\textbf{CW-ERM}) starts with the selection of a coreset, a representative subset of the original dataset, to focus the computationally intensive reweighting process. Upon fine-tuning these weights, we then broadcast them back to the entire dataset, thereby achieving a more globally effective reweighting. This unified process aims to balance both computational efficiency and model performance, streamlining the path from data selection to final model training.

\subsection{Coreset Selection}

Our method begins by leveraging pre-trained models benchmarked on ImageNet as feature extractors to extract descriptive features from the dataset for the following coreset selction. This provides a robust foundation for the subsequent processes.
Given a dataset $\mathcal{S} = \{(\mathbf{x}_1, y_1), \ldots, (\mathbf{x}_n, y_n)\}$, we utilize a pretrained model $h$ to extract a feature vector $\mathbf{z}_i = h(\mathbf{x}_i)$ for each data point $\mathbf{x}_i$. We also compute the median feature vector $\Bar{\mathbf{z}}_{y_i}$ for each class label $y_i$.
Armed with these feature representations, we proceed to select a compact yet representative coreset. Specifically, we employ the state-of-the-art moderate coreset algorithm \cite{xia2022moderate} for coreset selection. This technique works by calculating the distance $d(s_i) = ||\mathbf{z}_i - \mathbf{z}_{y_i}||_2$ between each data point $\mathbf{z}_i$ and its corresponding class median feature $\mathbf{z}_{y_i}$. Data points with distances closest to the median distance are selected for the coreset.
In summary, we first extract expressive features using pre-trained models and then leverage these features to select a representative coreset via median distance-based scoring. This provides an informative subset for subsequent processing.
We opted for moderate coreset as our coreset selection strategy due to its ability to provide a representative subset without expensive optimization procedures. By selecting data points close to the median distance, moderate coreset offers an efficient way to summarize the dataset while reducing the influence of outliers.

\subsection{Coreset Reweighting}
Following the coreset selection, we transition to the reweighting phase. In this stage, we employ MetaWeightNet~\cite{shu2019meta}, a well-regarded reweighting algorithm to optimize the sample weight of coreset \( \mathbf w^c \). Applying MetaWeightNet exclusively to the coreset alleviates the computational intensity usually associated with this method, striking a balance between efficiency and effectiveness. The reweighting procedure revolves around two loss functions: the weighted training loss, \(L_{\text{train}}\), and the meta loss, \(L_{\text{meta}}\). Our aim is to minimize \(L_{\text{train}}\) concerning the classifier weights \(\boldsymbol  \theta \), and \(L_{\text{meta}}\) with respect to the meta-parameters \(\boldsymbol \Theta\) in the weight network \(V(L; \boldsymbol \Theta)\).
The optimization is an iterative two-step process. Initially, \(L_{\text{train}}\) is minimized by updating \(\boldsymbol \theta\) in line with its gradient \(\nabla_{\boldsymbol \theta} L_{\text{train}}\). Subsequently, \(L_{\text{meta}}\) is optimized by altering \(\boldsymbol \Theta\) based on its gradient \(\nabla_{\boldsymbol \Theta} L_{\text{meta}}\). This approach allows for the joint optimization of \(\boldsymbol \theta\) and \(\boldsymbol \Theta\), contributing to the efficacy of our reweighting method. Employing MetaWeightNet justifies itself by its proven ability to learn a suitable weighting function through the concurrent optimization of both the training and meta objectives. Moreover, the selective use of a coreset for reweighting substantially bolsters computational efficiency, enabling us to capture the advantages of reweighting without the usually high computational costs.

\begin{algorithm}[t]
\caption{Classification via Coreset Reweighting}
\begin{algorithmic}[1]

\State \textbf{Input:} Full dataset \( \mathcal{S} \), Pretrained model \( h(\cdot) \)
\State \textbf{Output:} Trained model \( f \)

\State \textbf{Feature Extraction:} \( \mathbf{z} = h(\mathbf{x}) \)

\State \textbf{Coreset Selection:} \( \mathcal{S}^* \gets \text{ModerateCoreset}(\mathcal{S}, \mathbf{z}) \)

\State \textbf{Coreset Reweighting:} Initialize \( w, \Theta \)
\State \quad Optimize \( w \) and \( \Theta \) until convergence \cite{shu2019meta}

\State \textbf{Weight Broadcasting:} \( W_{S}(\mathbf{x}) \gets W_{{S}^*}(\text{NN}(\mathbf{x}, \mathcal{S}^*)) \) for all \( \mathbf{x} \in \mathcal{S} \)

\State \textbf{Model Training:} Initialize \( f \)
\State \quad Update \( f \) using \( \mathcal{S} \) and \( W_{S} \) until convergence

\end{algorithmic}
\end{algorithm}

\subsection{Weight Broadcasting}
The final step in our proposed methodology is the weight broadcasting phase, which aims to propagate the optimized weights \( \mathbf w^c \) from the coreset \( \mathcal{C} \) back to the full dataset \( \mathcal{S} \). The crux of this operation is captured by the \textit{broadcast} function, formulated as follows:
\begin{equation}
\mathbf w^*_i = \text{broadcast}(\mathbf{x}_i, \mathcal{C}, \mathbf w^c) = \mathbf  w^c_{\text{NN}(\mathbf{x}_i, \mathcal{C})},
\end{equation}
\label{eq: weight_broadcast}
where \( \text{NN}(x_i, \mathcal{C}) \) represents the index of the nearest neighbor of \( \mathbf{x}_i \) within the coreset \( \mathcal{C} \), and \( \mathbf  w^c_{\text{NN}(\mathbf{x}_i, \mathcal{C})} \) is the weight of that nearest neighbor in the coreset.
We employ a nearest neighbors algorithm to find the closest sample in the coreset for each sample in the original dataset. This ensures that the weights from the coreset are broadcasted to similar samples in \( \mathcal{S} \), effectively capturing the underlying distribution and characteristics of the data. The nearest neighbors method allows us to perform this broadcasting in a computationally efficient manner while maintaining the effectiveness of the reweighting, thus contributing to enhanced model performance across the entire dataset.

\definecolor{graycell}{rgb}{0.88,0.88,0.88}

\section{Experimental Results}
\label{sec:experiments}

\begin{table*}[t]
\centering
\resizebox{.84\linewidth}{!}{%
\begin{tabular}{cl|cc|cc|cc|cc}
    \toprule
    & \textbf{Classifier}: & \multicolumn{4}{c|}{PreactResnet18} & \multicolumn{4}{c}{Resnet20} \\
    \midrule
    & \textbf{Feature Extractor}: & \multicolumn{2}{c|}{ViT} & \multicolumn{2}{c|}{Resnet50} & \multicolumn{2}{c|}{ViT} & \multicolumn{2}{c}{Resnet50} \\
    \midrule
     & \textbf{Metrics} & Acc.(\%) & Time(Min) & Acc.(\%) & Time(Min) & Acc.(\%) & Time(Min) & Acc.(\%) & Time(Min) \\
    \midrule
    \multicolumn{10}{c}{\cellcolor{graycell}CIFAR-10} \\
    \midrule
    \multirow{5}{*}{\rotatebox[origin=c]{90}{\textbf{\fontsize{10pt}{12pt}\selectfont Method}}}
    & \( \textit{ERM} \) & $94.1 \pm 0.2$ &$30.90 \pm 4.26$ & $93.6 \pm 0.3$ & $28.02 \pm 3.42$ & $92.2 \pm 0.1$ & $21.00 \pm 0.36$ & $92.3 \pm 0.3$ & $20.52 \pm 1.26$ \\
    & \( \textit{W-ERM} \) & $93.8 \pm 0.1$ & $308.70 \pm 9.24$ & $94.1 \pm 0.3$ & $304.68 \pm 2.76$ & $92.3 \pm 0.1$ & $281.64 \pm 5.70$ & $92.3 \pm 0.1$ & $281.22 \pm 7.86$ \\
    & \( C_{\text{R}}\textit{-ERM} \) & $40.6 \pm 2.3$ & $14.70 \pm 0.06$ & $40.2 \pm 1.2$ & $14.58 \pm 0.66$ & $40.6 \pm 0.9$ & $14.76 \pm 0.30$ & $40.4 \pm 1.4$ & $14.64 \pm 0.06$ \\
    & \( C_{\text{MS}}\textit{-ERM} \) & $40.1 \pm 0.5$ & $20.58 \pm 0.24$ & $39.9 \pm 0.9$ & $17.70 \pm 0.78$ & $39.9 \pm 0.3$ & $19.92 \pm 0.72$ & $39.4 \pm 0.6$ & $17.46 \pm 0.30$ \\
    & \( \textbf{\textit{CW-ERM(Ours)}} \) & $\mathbf{94.9 \pm 0.1}$ &$54.66 \pm 0.84$ & $\mathbf{95.0 \pm 0.3}$ & $54.54 \pm 1.08$ & $\mathbf{92.5 \pm 0.1}$ & $32.40 \pm 1.44$ & $\mathbf{92.4 \pm 0.1}$ & $29.70 \pm 0.48$ \\
    \midrule
    \multicolumn{10}{c}{\cellcolor{graycell}CIFAR-100} \\
    \midrule
    \multirow{5}{*}{\rotatebox[origin=c]{90}{\textbf{\fontsize{10pt}{12pt}\selectfont Method}}}
    & \( \textit{ERM} \) & $76.2 \pm 0.2$ & $2940.00 \pm 3.36$ & $76.0 \pm 0.3$ & $28.02 \pm 2.88$ & $68.2 \pm 0.6$ & $22.68 \pm 0.72$ & $68.3 \pm 0.2$ & $22.80 \pm 1.92$ \\
    & \( \textit{W-ERM} \) & $76.3 \pm 0.3$ & $308.46 \pm 2.70$ & $76.3 \pm 0.2$ & $307.80 \pm 3.24$ & $68.1 \pm 0.4$ & $281.76 \pm 3.12$ & $68.2 \pm 0.1$ & $280.86 \pm 0.72$ \\
    & \( C_{\text{R}}\textit{-ERM} \) & $8.3 \pm 0.3$ & $14.70 \pm 0.06$ & $8.3 \pm 0.3$ & $15.00 \pm 0.48$ & $7.8 \pm 0.6$ & $17.16 \pm 0.66$ & $7.8 \pm 0.6$ & $15.24 \pm 0.60$ \\
    & \( C_{\text{MS}}\textit{-ERM} \) & $8.6 \pm 0.2$ & $21.72 \pm 0.72$ & $9.4 \pm 0.3$ & $18.24 \pm 0.66$ & $7.8 \pm 0.5$ & $20.58 \pm 0.12$ & $8.8 \pm 0.6$ & $17.70 \pm 0.78$ \\
    & \( \textbf{\textit{CW-ERM(Ours)}} \) & $\mathbf{76.7 \pm 0.2}$ & $57.30 \pm 1.08$ & $\mathbf{76.5 \pm 0.4}$ & $54.60 \pm 1.74$ & $\mathbf{68.5 \pm 0.2}$ & $33.36 \pm 0.30$ & $\mathbf{68.5 \pm 0.5}$ & $29.88 \pm 2.10$ \\
    \bottomrule
\end{tabular}%
}
\caption{\footnotesize{Experimental results comparing test accuracy and training time across various methods on the CIFAR-10 and CIFAR-100 datasets. Results represent averages from five distinct random seed trials. The methods compared encompass \textit{ERM}, \textit{W-ERM}, \(C_{\text{R}}\)-\textit{ERM}, \(C_{\text{MS}}\)-\textit{ERM}, and our proposed approach \textit{CW-ERM}. Experiments are conducted with two classifier backbones (PreactResnet18 and Resnet20) and two feature extractors (ViT and Resnet50). It should be noted that, for consistency, the Coreset Ratio for all methods is fixed at 0.01. Methods \(C_{\text{R}}\)-\textit{ERM} and \(C_{\text{MS}}\)-\textit{ERM} employ random and moderate coreset selection, respectively, and exclusively train on the coreset without applying broadcasting. Unlike them, \textit{ERM} and \textit{W-ERM} do not utilize coresets.}}
\label{tab:results}
\end{table*}
\vspace{-4mm}

\textbf{Experiment setup.} 
We assess the efficacy of our proposed method, \textbf{CW-ERM}, on the CIFAR-10 and CIFAR-100 datasets \cite{krizhevsky2009learning}. These datasets are benchmarks in the machine learning community and offer a robust platform for comparing our method against established baselines such as ERM, Weighted-ERM (\textbf{W-ERM}), Coreset-Random-ERM (\(\textbf{C}_{\text{R}}\)-\textbf{ERM}), and Coreset-Moderate-ERM (\(\textbf{C}_{\text{MS}}\)-\textbf{ERM})
. The training is carried out using the PreactResNet18 architecture \cite{he2016identity} and optimized using Stochastic Gradient Descent (SGD \cite{ruder2016overview}) with a learning rate of 0.1, momentum of 0.9, and a weight decay of \(5 \times 10^{-4}\). For feature extraction, we utilize a pre-trained Vision Transformer (ViT \cite{dosovitskiy2021image}). 
\begin{figure}[thb]
  \begin{center}
    \begin{subfigure}[b]{0.23\textwidth} 
      \includegraphics[width=\textwidth]{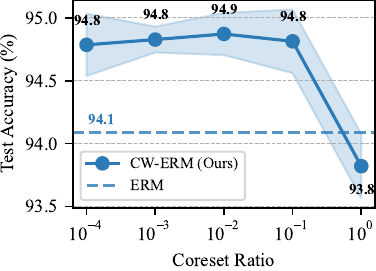} 
      \vspace*{-1.0em}
      \caption{CIFAR-10}
      \label{fig:coreset_ratios_cifar10}
    \end{subfigure}
    \begin{subfigure}[b]{0.23\textwidth} 
      \includegraphics[width=\textwidth]{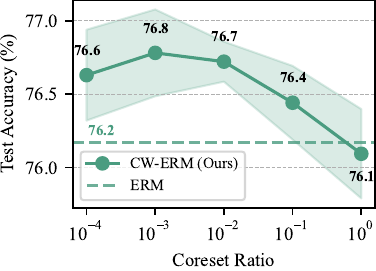} 
      \vspace*{-1.0em}
      \caption{CIFAR-100}
      \label{fig:coreset_ratios_cifar100}
    \end{subfigure}
    \vspace*{-1.0em}
    \caption{\footnotesize{The effect of the coreset ratio on the model performance on CIFAR-10 and CIFAR-100, where coreset is solely used for data reweighting. After reweighting data of the coreset, weights are broadcasted back to the full dataset for training. Larger coreset ratios may lead to test accuracy degradation, particularly in more complex datasets like CIFAR-100. "Uniform data reweighting" refers to the process where each data point in the dataset is treated with equal importance, without any specialized weighting scheme.}}
    \label{fig:coreset_ratios}
  \end{center}
  \vspace*{-1.5em}
\end{figure}

\begin{wrapfigure}{r}{48mm}
\hspace{-3mm}
\vspace*{-4mm}
\centerline{
\includegraphics[width=48mm,height=!]{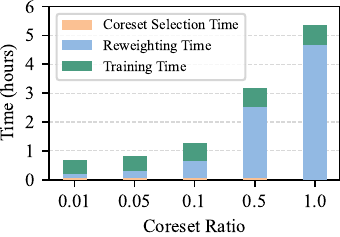}
}
\caption{\footnotesize{
Time consumption breakdown for 5 different coreset ratios. The stacked bar chart shows the average time spent on coreset selection, reweighting, and training for each coreset ratio. 
}}
\label{fig:time-comparison}
\vspace{-5mm}
\end{wrapfigure}

\noindent\textbf{Overall performance.}
As detailed in \textbf{Table \ref{tab:results}}, our proposed \textbf{CW-ERM} consistently exceeds the performance of other methods in both CIFAR-10 and CIFAR-100 datasets, especially noteworthy at a coreset ratio of 0.01 where it achieves average accuracies of 94.9\% and 76.7\%
, respectively. 
This superior accuracy is further substantiated in \textbf{Fig. \ref{fig:coreset_ratios}}, which provides a nuanced exploration of the trade-offs between accuracy and computational efficiency. Here, \textbf{CW-ERM} not only achieves superior classification but also reduces computational time. At a coreset ratio of 0.01, it realizes a mean accuracy of 94.9\% with a standard deviation of 0.1\%, significantly surpassing the baseline accuracy of 94.1\% set by ERM. The computational benefits of \textbf{CW-ERM} are further highlighted in \textbf{Fig. \ref{fig:time-comparison}}, which delineates the time consumption for various phases of the learning process across 5 different coreset ratios. The method significantly curtails the time spent in the computationally intensive reweighting phase without sacrificing accuracy, further corroborating the method's robustness and efficiency.
Our experiments substantiate that CW-ERM offers a robust and efficient strategy for machine learning tasks. The method demonstrates significant improvements in both classification accuracy and computational efficiency, thereby providing a compelling case for its applicability in real-world scenarios.

\section{Conclusion}
\label{sec:conclusion}

In this work, we have presented a novel method for enhancing the efficiency and accuracy of machine learning classification tasks. Our approach combines the principles of coreset selection and data reweighting, delivering a system that outperforms existing techniques in terms of computational time and model performance. By conducting experiments on widely acknowledged datasets like CIFAR-10 and CIFAR-100, we have demonstrated the method's scalability and effectiveness.
Our results confirm that it is possible to achieve a delicate balance between computational efficiency and model accuracy, a challenge that has long plagued the machine learning community. The promising results from this study pave the way for future research in this direction, including potential extensions to other types of learning tasks and datasets. We believe that our approach serves as a strong foundation for developing scalable and accurate machine learning systems.

\vspace{2mm}
\noindent \textbf{Acknowledgement} 
The work of Yimeng Zhang, Yihua Zhang, and Sijia Liu was supported by the Cisco Research Award.

\newpage

\bibliographystyle{IEEEbib}
\bibliography{refs}

\end{document}